\documentclass[opre,nonblindrev]{informs3}

\DoubleSpacedXI 

\usepackage{endnotes}
\let\footnote=\endnote
\let\enotesize=\normalsize
\def\notesname{Endnotes}%
\def\makeenmark{\hbox to1.275em{\theenmark.\enskip\hss}}
\def\enoteformat{\rightskip0pt\leftskip0pt\parindent=1.275em
  \leavevmode\llap{\makeenmark}}

\usepackage{algorithm}
\usepackage{algpseudocode}
\usepackage{graphicx}
\usepackage{bbm}
\usepackage{amsfonts}
\usepackage{amsmath}

\usepackage{natbib}
 \bibpunct[, ]{(}{)}{,}{a}{}{,}%
 \def\bibfont{\small}%
\TheoremsNumberedThrough     
\ECRepeatTheorems

\EquationsNumberedThrough    

\begin{document}


\RUNAUTHOR{Patil and Mintz}

\RUNTITLE{A Mixed-Integer Programming Approach to Training Dense Neural Networks}

\TITLE{A Mixed-Integer Programming Approach to Training Dense Neural Networks}

\ARTICLEAUTHORS{%
\AUTHOR{Vrishabh Patil, Yonatan Mintz}
\AFF{Department of Industrial and Systems Engineering, University of Wisconsin-Madison, Madison, WI 53706, USA \EMAIL{vmpatil@wisc.edu}, \EMAIL{ymintz@wisc.edu}} 
} 

\ABSTRACT{%
Artificial Neural Networks (ANNs) are prevalent machine learning models that are applied across various real world classification tasks. However, training ANNs is time-consuming and the resulting models take a lot of memory to deploy. In order to train more parsimonious ANNs, we propose a novel mixed-integer programming (MIP) formulation for training fully-connected ANNs. Our formulations can account for both binary and rectified linear unit (ReLU) activations, and for the use of a log-likelihood loss. We present numerical experiments comparing our MIP-based methods against existing approaches and show that we are able to achieve competitive out-of-sample performance with more parsimonious models.
}%



\maketitle

%


%
\section{Introduction}
Artificial Neural Networks (ANNs) have become ubiquitous across numerous classification tasks in domain areas such as natural language processing and image recognition. Typically, these existing models are trained through the use of end-to-end back propagation and stochastic gradient descent (SGD). These methods attempt to minimize a global loss function which is a function of the network architecture and data by taking steps in the direction of a negative gradient that is calculated incrementally using the chain rule \citep{goodfellow2017deep}. While these end-to-end methods have been shown to have good empirical performance \citep{krizhevsky2012imagenet,szegedy2015going} they suffer from several well known limitations. For example, there are several computational challenges with end-to-end methods such as the vanishing gradient problem (where the gradient computed through back propagation fails to update the parameters for large network architectures) \citep{bengio1994learning, hochreiter2001gradient}, and the numerical convergence of SGD is sensitive to random initialization \citep{bottou1991stochastic, sutskever2013importance}. However, a major shortcoming that underlies these computational challenges is that to obtain strong out of sample performance, models using end-to-end methods often times need to have complex network architectures and host numerous parameters while training. This means that training these models often costly with regards to time and memory \citep{taylor2016training}. Given this challenge, recent research has explored if more parsimonious and stable ANNs can be trained without using end-to-end training methods \citep{elad2018effectiveness,lowe2019putting,belilovsky2019greedy}. 
In this paper, we continue to explore this line of inquiry, by proposing none end-to-end ANN training methods based on principles from mixed-integer programming (MIP) and constrained optimization. In particular we present MIP formulations for network architectures that contain either binary or rectified linear unit (ReLU) activation functions. In addition our contributions include a MIP formulation for the negative log likelihood cost in multi-class classification, making our formulations compatible with common soft-max implementations. We also discuss how our formulations can be used in a layer by layer training approach which can be used effectively as a pre-training procedure for large network architectures. Finally, we present computational results of how our methods compare to SGD-based training methods in terms of resulting model size and performance.

\section{Related Works}
In this section, we present some related streams of literature that have explored the intersection of ANNs and MIP, and discuss how they relate to our work. We discuss some work related to training ANNs using MIPs, followed by papers that focus on the convexification of the training problem, along with studies involving the amalgamation of MIPs with ANNs. Finally, we review methodological papers that relate to greedy layer-wise algorithms in machine learning. 

In recent years, MIPs have gained popularity in training neural networks. Constraint Programming, MIP, and a hybrid of the two have been used to train Binarized Neural Networks (BNNs), a class of neural networks with the weights and activations restricted to the set \{-1, +1\}. In \cite{toro2019training} the authors report on the availability of an optimal solution at an acceptable optimality gap, given limitation on the size of the training set. To relax the binarized restriction on weights, \cite{thorbjarnarsontraining} propose a formulation to train Integer Neural Networks with different loss functions. However, all the experiments in the paper were restricted to 100 samples. Continuing along these lines, \cite{kurtz2021efficient} has also develop a formulation for integer weights, and offer an iterative data splitting algorithm that uses the \emph{k-means} classification method to train on subsets of data with similar activation patterns.

In a different stream, \cite{askari2018lifted} propose using convex relaxations for training ANNs. The authors also explore non gradient-based approaches and initialized weights to accelerate convergence of gradient-based algorithms. MIP formulations have been used for already trained networks to provide adversarial samples that can improve network stability \citep{fischetti2017deep,anderson2020strong, tjeng2017evaluating}. Finally, there has been research directed towards using ANNs to solve MIPs \citep{nair2020solving}. In contrast to the previous literature, we propose methods that directly use MIP techniques for training ANNs as opposed to only being used in their evaluation.

With regards to pre-training neural networks using greedy layer-wise algorithms, models introduced in \cite{bengio2007greedy} have been explored actively in the machine learning literature
\citep{erhan2010does, belilovsky2019greedy, elad2018effectiveness}. The \emph{Greedy Infomax} algorithm \citep{lowe2019putting} has also been shown to have good performance in training ANNs without the need for end-to-end back-propagation. These results indicate that training deep networks through decoupled training can still result in strong out of sample performance.

Our contributions to this field of literature are threefold: we propose a tight MIP formulation to train ANNs without integrality constraints on the weight matrix and develop a novel approach that addresses nonlinear activations. We also formulate a negative log-likelihood loss that allows for a soft-max output. Finally, we further contribute to literature pertaining the greedy layer-wise algorithms by showing that the use of MIP in solving the layer by layer training problems can result in more parsimonious ANN models that achieve competitive out of sample performance.

\section{MIP Formulations for ANN Training}
Suppose we have a data set composed of ordered pairs $(x_n,y_n)$ for $n = 1,...,N$ where $x_n \in \mathcal{X} \subset \mathbbm{R}^d$ is the feature vector of data point $n$ and $y_n \in J \subset \mathbbm{Z}$ is its label. We use the notation $[N]_m = \{m,m+1,...,N\}$ for any two integers $N>m$ and assume that $|J| < \infty$. Our goal is to find a function $f: \mathbbm{R}^d \mapsto J$ that is parametrized by $\theta \in \Theta$ such that for any $x\in \mathcal{X}$, $f(x;\theta)$ is able to closely predict the value of the corresponding $y$. This $f$ is obtained by finding $\hat{\theta} \in \Theta$ that minimizes the empirical negative log likelihood loss $\mathcal{L}:J \times J \mapsto \mathbbm{R}$ that is $\hat{\theta} = \mathrm{argmin}_{\theta \in \Theta} \sum_{n = 1}^N \mathcal{L}(f(x,\theta),y)$. We assume $f$ is an ANN that can be expressed as a functional composition of $L$ vector valued functions that is $f(x;\theta) = h_L\circ h_{L-1} \circ ... \circ h_1 \circ h_0(x)$. Here each function $h_\ell(\cdot,\theta_\ell)$ is referred to as a layer of the neural network, we assume that $h_0:\mathcal{X} \mapsto R^K$, $h_L:\mathbbm{R}^K \mapsto J$, and $h_\ell: \mathbbm{R}^K \mapsto \mathbbm{R}^K$, we denote each component of $h_\ell$ as $h_{k,\ell}$ for $\ell \in [L-1]_0$ and $k \in [K]_1$, and each component of $h_L$ as $h_{j,L}$ for $j \in [J]_1$. We refer to these components as units or neurons. We refer to the index $\ell$ as the layer number of the ANN, and the index $k$ as the unit number. The layer corresponding to the index $L$ is referred to as the output layer, and all other indices are referred to as hidden layers. Each unit has the functional form: $h_{k,0} = \sigma(\alpha_{k,0}^\top x + \beta_{k,0}) $, $h_{k,\ell} = \sigma(\alpha_{k,\ell}^\top h_{\ell-1} + \beta_{k,\ell})$ for $\ell \in [L-1]_1$, $h_{j,L} = \varphi(\alpha_{j,L}^\top h_{L-1} + \beta_{j,L})$, where $\sigma:\mathbbm{R}\mapsto \mathbbm{R}$ is a non-linear function applied over the hidden layers $\ell = 0,...,L-1$ called the activation function, $\varphi:\mathbbm{R}\mapsto \mathbbm{R}$ is a different non-linear function applied over the output layer $L$, and $ \alpha $ and $ \beta $ are the weights and biases matrices respectively where $(\alpha,\beta) = \theta$. The notation $\alpha_{a,b,\ell}$ indicates the weight being applied to unit index $a$ in layer $\ell -1$ for the evaluation of unit index $b$ in layer $\ell$. Likewise, $\beta_{k,\ell}$ indicates the bias associated with unit index $k$ in layer $\ell$ of the network. We consider two potential activation forms, either binary or ReLU activation. If $\sigma$ is the binary activation then we assume for any input $z \in \mathbbm{R}$, $\sigma(z) = \mathbbm{1}[z \geq 0]$, where $\mathbbm{1}$ is the indicator function. If $\sigma$ is a ReLU activation function then $\sigma(z) = \max\{z,0\}$. 

Our goal is to show how the above training problem and model can be formulated as a MIP. The key to our reformulation is the introduction of decision variables $h_{n,k,\ell}$ which correspond to the unit output of unit $k$ in layer $\ell$ when the ANN is evaluated at data point with index $n$. Having these decision variables be data point dependant, and ensuring that $\alpha_{k^\prime,k,\ell},\beta_{k,\ell}$ are the same across all data points forms the back bone of our formulation. Thus, if $x_{n,i}$ denote the $i^{th}$ feature value of data point $n$, the general form of the optimization problem is:
\begin{flalign}
\min & \sum_{n=1}^N\sum_{j \in J} \mathcal{L}(h_{n,j,L},y_{n,j}) & \\
\textrm{subject to } \notag\\
 & h_{n,k,0} = \sigma(\sum_{i=1}^d\alpha_{i,k,0}x_{n,i} + \beta_{k,0}), \ \forall \ k,n \in [K]_1 \times [N]_1 \label{l_0 output_const}  \\
 & h_{n,k,\ell} = \sigma(\sum_{k'=1}^K\alpha_{k^\prime,k,\ell}h_{n,k^\prime,\ell-1} + \beta_{k,\ell}), \ \forall \ \ell,k,n \in [L-1]_1 \times [K]_1, \times [N]_1 \label{l output_const} \\
 & h_{n,j,L} = \varphi(\sum_{k'=1}^K\alpha_{k^\prime,j,L}h_{n,k^\prime,L-1} + \beta_{j,L}), \ \forall j,n \in |J| \times [N]_1 \label{L output_const} \\
 & \alpha_{i,k,0},\beta_{k,0},\alpha_{k^\prime,k,\ell}, \beta_{k,\ell}, \alpha_{k^\prime,j,L},\beta_{j,L} \in \Theta, \ \forall \ i,\ell,k,k^\prime,j \in [d]_1 \times [L]_1 \times [K]_1^2\times |J|
\end{flalign}

\subsection{MIP Formulation of Loss Function\label{NLL}}
First we focus on the reformulation of the objective function $\sum_{n=1}^N\sum_{j \in J} \mathcal{L}(h_{n,j,L},y_{n,j})$ in terms of a set of linear constraints and linear objective function. Note that the the $ L^{th} $ layer outputs are the only ones that interact directly with the loss function, so we focus on these decision variables and leave the discussion of the model parameters when considering individual unit activations. Objectives such as prediction inaccuracy and $\ell_0$ or $\ell_1$ losses can be trivially formulated using linear and MIP techniques. However, in practice a common loss function used for training ANNs in classification tasks, is the negative log likelihood applied to a soft-max distribution \citep{goodfellow2017deep}. This kind of loss can be considered analogous to minimizing the log likelihood of a multinomial distribution across the classes in $J$, thus outputting a predictive distribution over the classes. This loss is particularly appealing as it harshly penalizes incorrect classifications and is numerically stable \citep{goodfellow2017deep}. The equation for the soft-max activation function for the output of the final layer $ h_{n,j,L} $ is $\varphi(h_{n,j,L}) = \frac{\exp(h_{n,j,L})}{\sum_{j' \in J} \exp(h_{n,j',L})} \label{softmax}$, and the resulting negative log likelihood loss is $-\log(\prod_{n=1}^{N}(\varphi(h_{n,j,L})^{\mathbbm{1}[y_n=j]}))$.

Our main result is:
\begin{proposition}
The negative log likelihood loss applied to the soft-max activation $-\log(\prod_{n=1}^{N}(\varphi(h_{n,j,L})^{\mathbbm{1}[y_n=j]}))$ is within a constant additive error of the optimal value of the following linear optimization problem:
\begin{align}
 \min_{h_{n,j,L},\omega_n}\big\{ \sum_{n=1}^{N} \sum_{j=1}^{J} \mathbbm{1}[y_n=j](\omega_{n}-h_{n,j,L}) : \omega_n \geq h_{n,j^\prime,L}, \ j^\prime\in [J
]_1,n \in [N]_1 \big\} \label{obj}
\end{align}
\label{prop:nll}
\end{proposition}
Here we present a brief sketch of the proof, for the detailed proof of this proposition please see the appendix. The main insight used is the fact that the function $\log \sum_{i=1}^n \exp(x_i) = \Theta(\max_{i \in [N]_0} x_i)$ \citep{calafiore2014optimization}, where $\Theta$ is the Big Theta notion that characterizes the function $\log \sum\exp$ to be bounded asymptotically both above and below by the $max$ function \citep{cormen2022introduction}, and the fact that the minimization of a maximum can be written using linear constraints \citep{wolsey1999integer}.

\subsection{MIP Formulation for Binary Activated ANNs\label{Binary MIP}}

In this section, we present a MIP reformulation for ANN units with binary activation that can be solved using commercial solvers. We rewrite Constraints \eqref{l_0 output_const},\eqref{l output_const},\eqref{L output_const} for a single unit as $h_{n,k,0} = \mathbbm{1}[ \sum_{i =1}^d \alpha_{i,k,0}x_{i} + \beta_{k,0} \geq 0]$, $h_{n,k,\ell} = \mathbbm{1}[ \sum_{k^\prime =1}^K \alpha_{k,k^\prime,\ell}h_{k^\prime,\ell} + \beta_{k,\ell} \geq 0]$, $h_{n,j,L} = \sum_{k^\prime =1}^K \alpha_{k^\prime,j,L}h_{k^\prime,L-1} + \beta_{j,\ell} $ respectively. Note that for all layers that are not the input layer, the above constraints contain bi-linear products which make this formulation challenging to solve. As such we propose the following reformulation:

\begin{proposition}
 In the binary activation case, the Constraints \eqref{l_0 output_const},\eqref{l output_const},\eqref{L output_const} can be reformulated as a set of MIP constraints. Specifically for all $k\in [K]_1,n\in[N]_1$ Constraint \eqref{l_0 output_const} can be reformulated as:
 \begin{align}
 & \sum_{i=1}^{d} (\alpha_{i,k,0}x_{n,i}) + \beta_{k,0} \le Mh_{n,k,0} \label{h0_define C1} \\
 & \sum_{i=1}^{d} (\alpha_{i,k,0}x_{n,i}) + \beta_{k,0} \ge \epsilon + (-M-\epsilon)(1-h_{n,k,0}) \label{h0_define C2} 
 \end{align}
 For all $\ell \in [L-1]_1,k\in[K]_1,n\in[N]_1$ Constraints \eqref{l output_const} can be reformulated as:
\begin{align}
 & \sum_{k^\prime=1}^{K} (z_{n,k^\prime,k,\ell}) + \beta_{k,\ell} \le Mh_{n,k,\ell} \label{hl_define C1} \\
 & \sum_{k^\prime=1}^{K} (z_{n,k^\prime,k,\ell}) + \beta_{k,\ell} \ge \epsilon + (-M-\epsilon)(1-h_{n,k,\ell}) \label{hl_define C2} \\
 & z_{n,k^\prime,k,\ell} \le \alpha_{k^\prime,k,\ell} + M(1-h_{n,k^\prime,\ell-1}), \ \forall \ k^\prime \in [K]_1 \label{zl_define C1}\\
 & z_{n,k^\prime,k,\ell} \ge \alpha_{k^\prime,k,\ell} -M(1-h_{n,k^\prime,\ell-1}), \ \forall \ k^\prime \in [K]_1 \label{zl_define C2}\\ 
 & -Mh_{n,k^\prime,\ell-1} \le z_{n,k^\prime,k,\ell} \le Mh_{n,k^\prime,\ell-1}, \ \forall \ k^\prime \in [K]_1 \label{zl_define C3}
\end{align}
And for all $j\in J,n \in [N]_1$, Constraints \eqref{L output_const} can be reformulated as:
\begin{align}
 & h_{n,j,L} \le \sum_{k^\prime=1}^{K} (z_{n,k^\prime,j,L}) + \beta_{j,L} \label{hL_define C1} \\
 & h_{n,j,L} \ge \sum_{k^\prime=1}^{K} (z_{n,k^\prime,j,L}) + \beta_{j,L} \label{hL_define C2} \\
 & z_{n,k^\prime,j,L} \le \alpha_{k^\prime,j,L} + M(1-h_{n,k^\prime,L-1}), \ \forall \ k^\prime \in [K]_1 \label{zL_define C1} \\
 & z_{n,k^\prime,j,L} \ge \alpha_{k^\prime,j,L} - M(1-h_{n,k^\prime,L-1}), \ \forall \ k^\prime \in [K]_1 \label{zL_define C2} \\
 & -Mh_{n,k^\prime,L-1} \le z_{n,k^\prime,j,L} \le Mh_{n,k^\prime,L-1}, \ \forall \ k^\prime \in [K]_1 \label{zL_define C3} \\
 & \omega_{n} \ge h_{n,j,L} \label{omega_define} \\
 & h_{n,j,L} + h_{n,j^\prime,L} - 2h_{n,j,L} \leq - \epsilon + Mr_{n,j,j^\prime}, \ \forall \ j,j^\prime \in [J]^{2}_{1}, j \neq j^\prime \label{diversify C1} \\
 & h_{n,j,L} + h_{n,j^\prime,L} - 2h_{n,j,L} \geq \epsilon - M(1-r_{n,j,j^\prime}), \ \forall \ j,j^\prime \in [J]^{2}_{1}, j \neq j^\prime \label{diversify C2} 
\end{align}
\label{prop:reform_bin}
 \end{proposition}

Constraints \eqref{h0_define C1} and \eqref{h0_define C2} are big M constraints that, when combined with the integrality of $h_{n,k,0}$, impose the output of the first hidden layer to be 1 if the linear combination of the units of the input vector summed with the bias term is greater than or equal to some small constant $\epsilon$, and 0 otherwise. To reformulate Constraints \eqref{l output_const}, we introduce an auxiliary variable $z_{n,k^\prime,k,\ell} = \alpha_{k^\prime,k,\ell}h_{n,k^\prime,\ell-1}$. Constraints \eqref{hl_define C1} and \eqref{hl_define C2} are then similar to Constraints \eqref{h0_define C1} and \eqref{h0_define C2}, and define the output of the remaining hidden layers $\ell \in [L-1]_{1}$. We leverage the fact that the bi-linear term is a product of a continuous and binary variable for Constraints \eqref{zl_define C1}, \eqref{zl_define C2}, and \eqref{zl_define C3};  $z_{n,k^\prime,k,\ell}=0$ when $h_{n,k^\prime,\ell-1}=0$ or $z_{n,k^\prime,k,\ell}=\alpha_{k^\prime,k,\ell}$ when $h_{n,k^\prime,\ell-1}=1$. A symmetric argument holds for the definition of Constraints \eqref{zL_define C1}, \eqref{zL_define C2}, and \eqref{zL_define C3}, which ensures $z_{n,k^\prime,j,L} = \alpha_{k^\prime,j,L}h_{n,k^\prime,L-1}$, the bi-linear terms associated with the output layer. Considering that the soft-max activation of the output layer is captured by the objective function as defined in Section \ref{NLL}, Constraints \eqref{hL_define C1} and \eqref{hL_define C2} guarantee that the output of the final layer is the linear combination of the units of the activated outputs of the penultimate layer summed with the bias term. Constraints \eqref{omega_define} forces $\omega_{n}$ to be the \emph{max} of the components of the output vector. Finally, Constraints \eqref{diversify C1} and \eqref{diversify C2} ensure that no two units in the output layer are equal. The idea behind these two constraints depends on the fact that the average of two variables is equal to one of the variables if and only if the two variables are equal. As such, we introduce the binary variables $r_{n,j,j^\prime} \in \mathbbm{B}$ to define Constraints \eqref{diversify C1} and \eqref{diversify C2} as big M constraints.

A full proof of this proposition can be found in the appendix. However, we will present a sketch here. The main techniques for this proof rely on first using a big M formulation for disjunctive constraints \citep{wolsey1999integer} to model the binary activation. Then bi-linear terms are reformulated using the techniques applied to products of binary and continuous variables. We note that for the output of the final layer to give a unique prediction with the log likelihood soft-max loss, we add Constraints (\ref{diversify C1}) and (\ref{diversify C2}). We call these constraints diversifying constraints, which force the optimization problem to assign diverse values to $ h_{n,j,L}, \forall \; n \in [N]_{1},j \in J $. Without these constraints, we could technically minimize our loss by setting the output of the final layer, $ h_{n,j,L} $ to all be equal. However, owing to the fact that no data point can be labeled with multiple classes, we require unique unit outputs. The complete mathematical model for our problem can be found in the appendix.

\subsection{MIP Formulations for ReLU Activated ANNs\label{ReLU MIP}}

ANNs with ReLU activated hidden layer units are another class of models commonly used in practice. ReLU is commonly paired with soft-max loss functions as an activation function for training deep supervised ANNs \citep{goodfellow2017deep}. ReLU activations are especially beneficial in negating the vanishing gradient problem since the activation function has a gradient of 1 at all activated units \citep{glorot2011deep}. Although training ANNs with MIPs directly addresses the vanishing gradient problem by eliminating the need for back-propagation entirely, this issue can persist if the MIP model is used as pre-training initialization for an SGD approach to replace randomized initialization.
For this set of activations we can write Constraints \eqref{l_0 output_const},\eqref{l output_const}, for a single unit as $h^{ReLU}_{n,k,0} = \max\{ \sum_{i =1}^d \alpha_{i,k,0}x_{n,i} + \beta_{k,0}, 0 \}$, $h^{ReLU}_{n,k,\ell} = \max\{\sum_{k^\prime =1}^K \alpha_{k^\prime,k,\ell}h_{k^\prime,\ell} + \beta_{k,\ell}, 0 \}$ respectively. Similar to the binary case, the two main reformulation challenges that arise from these constraints are the piece-wise definition of the activation and the presence of bi-linear terms. Unlike the binary case however, the bi-linear terms involve the multiplication of two real valued decision variables and not a binary variable with a continuous variable. Since the resulting formulation would be challenging for commercial solvers to solve effectively we propose a relaxation formulation for the the ReLU activation case. In particular, we utilize piece-wise McCormick relaxations to reformulate the problem as a linear MIP.

\begin{proposition}
In the ReLU activation case, the Constraints \eqref{l_0 output_const},\eqref{l output_const} can be approximated using a MIP relaxation. Specifically, for a given partition number $P$, for all $k\in [K]_1, n\in [N]_1$:
\begin{align}
 & \sum_{i=1}^{d} (\alpha_{i,k,0}x_{n,i}) + \beta_{k,0} \le Mh_{n,k,0} \label{ReLU h0_define C1} \\
 & \sum_{i=1}^{d} (\alpha_{i,k,0}x_{n,i}) + \beta_{k,0} \ge \epsilon + (-M-\epsilon)(1-h_{n,k,0}) \label{ReLU h0_define C2} \\
 & h^{ReLU}_{n,k,0} \le (\sum_{i=1}^{D} (\alpha_{i,k,0}x_{n,d}) + \beta_{k,0}) + M(1-h_{n,k,0}) \label{ReLU h_relu0 C1} \\
 & h^{ReLU}_{n,k,0} \ge (\sum_{i=1}^{D} (\alpha_{i,k,0}x_{n,d}) + \beta_{k,0}) - M(1-h_{n,k,0}) \label{ReLU h_relu0 C2} \\
 & -Mh_{n,k,0} \le h^{ReLU}_{n,k,0} \le Mh_{n,k,0} \label{ReLU h_relu0 C3}
\end{align}

For all $p \in [P]_1,\ell \in [L-1]_1,k^\prime,k \in [K]^{2}_{1},n\in[N]_1$:
\begin{align}
 & z_{n,k^\prime,k,\ell} \geq \alpha_{p}^{L}h^{ReLU}_{n,k^\prime,\ell-1} - M(1-\lambda_{k^\prime,k,\ell,p}) \label{McCormick C1} \\
 & z_{n,k^\prime,k,\ell} \geq \alpha_{p}^{U}h^{ReLU}_{n,k^\prime,\ell-1} + \alpha_{k^\prime,k,\ell}h_{ReLU}^{U} - \alpha_{p}^{U}h_{ReLU}^{U} - M(1-\lambda_{k^\prime,k,\ell,p}) \label{McCormick C2} \\
 & z_{n,k^\prime,k,\ell} \leq \alpha_{p}^{U}h^{ReLU}_{n,k^\prime,\ell-1} + M(1-\lambda_{k^\prime,k,\ell,p}) \label{McCormick C3} \\
 & z_{n,k^\prime,k,\ell} \leq \alpha_{p}^{L}h^{ReLU}_{n,k^\prime,\ell-1} + \alpha_{k^\prime,k,\ell}h_{ReLU}^{U} - \alpha_{p}^{L}h_{ReLU}^{U} + M(1-\lambda_{k^\prime,k,\ell,p}) \label{McCormick C4}
\end{align} 

For all $\ell \in [L-1]_1,k^\prime,k \in [K]^{2}_{1}$
\begin{align}
 & \sum_{p=1}^{P} \lambda_{k^\prime,k,\ell,p} = 1 \label{McCormick C5} \\
 & \sum_{p=1}^{P}\alpha_{p}^{L}\lambda_{k^\prime,k,\ell,p} \le \alpha_{k^\prime,k,\ell} \le \sum_{p=1}^{P}\alpha_{p}^{U}\lambda_{k^\prime,k,\ell,p} \label{McCormick C6}
\end{align}

For all $ \ell \in [L-1]_1, k \in [K]_1, n \in [N]_1 $
\begin{align}
 & \sum_{k^\prime=1}^{K} (z_{n,k^\prime,k,\ell}) + \beta_{k,\ell} \le Mh_{n,k,\ell} \label{ReLU hl_define C1} \\
 & \sum_{k^\prime=1}^{K} (z_{n,k^\prime,k,\ell}) + \beta_{k,\ell} \ge \epsilon + (-M-\epsilon)(1-h_{n,k,\ell}) \label{ReLU hl_define C2} \\
 & h^{ReLU}_{n,k,\ell} \le (\sum_{k^\prime=1}^{K} (z_{n,k^\prime,k,\ell}) + \beta_{k,\ell}) + M(1-h_{n,k,\ell}) \label{ReLU h_relul C1} \\
 & h^{ReLU}_{n,k,\ell} \ge (\sum_{k^\prime=1}^{K} (z_{n,k^\prime,k,\ell}) + \beta_{k,\ell}) - M(1-h_{n,k,\ell}) \label{ReLU h_relul C2} \\
 & -Mh_{n,k,\ell} \le h^{ReLU}_{n,k,\ell} \le Mh_{n,k,\ell} \label{ReLU h_relul C3} 
\end{align}

Where $\alpha^{L}_{p} = \alpha^{L} + (\alpha^{U} - \alpha^{L}) \cdot (p-1)/P$ and $\alpha^{L}_{p} = \alpha^{L} + (\alpha^{U} - \alpha^{L}) \cdot p/P$
\label{prop:reform_relu}
\end{proposition}

A detailed proof of this proposition can be found in the appendix, however, we present a brief sketch here. Similar to Proposition \ref{prop:reform_bin}, we use big M formulations for disjunctive constraints \citep{wolsey1999integer} to obtain $h^{ReLU}_{n,k,\ell}$ from $h_{n,k,\ell}$, the binary indicator.

Constraints \eqref{ReLU h0_define C1} and \eqref{ReLU h0_define C2} are as defined in Section \ref{Binary MIP}. Constraints \eqref{ReLU h0_define C1} through \eqref{ReLU h_relu0 C3} then enforce $h^{ReLU}_{n,k,0} = \max\{ \sum_{i =1}^d \alpha_{i,k,0}x_{n,i} + \beta_{k,0}, 0\}$. Our formulation relies on piece-wise McCormick relaxations to reformulate the training problem as a linear MIP. To do so, we introduce an auxiliary variable $ z_{n,k^\prime,k,\ell} $ as a placeholder for $ \alpha_{k^\prime,k,\ell}h^{ReLU}_{n,k^\prime,\ell-1} $, partition $ \alpha $ into $ P $ pieces, and introduce a set of binary variables $\lambda_{n,k^\prime,k,\ell} \in \mathbbm{B}$. Partitioning on $\alpha$ gives $\alpha_{p}^{U}$ and $\alpha_{p}^{L}$, the respective upper- and lower-bounds on the partitions. The values for $ \alpha^{L}_{p} $ and $ \alpha^{U}_{p} $ are adapted from the partition bounds first defined by (\cite{castro2015tightening}). We also let $h_{ReLU}^{U}$ and $h_{ReLU}^{L}$ to be the respective upper- and lower-bounds on all $h^{ReLU}_{n,k^\prime,\ell}$. Constraints \eqref{McCormick C1} and \eqref{McCormick C2} act as under-estimators to $z_{n,k^\prime,k,l}$, while Constraints \eqref{McCormick C3} and \eqref{McCormick C4} act as over-estimators. Constraints \eqref{McCormick C5} chooses which partition best minimizes the objective. Constraints \eqref{McCormick C6} then enforces that $\alpha_{k^\prime,k,\ell}$ is bounded by the upper- and lower-bounds defined by the chosen partition. Given a large enough P, \citep{raman1994modelling} show that the McCormick relaxations are effective estimators for the bi-linear terms. Finally, Constraints \eqref{ReLU hl_define C1} through \eqref{ReLU h_relul C3} are analogous to Constraints \eqref{ReLU h0_define C1} through \eqref{ReLU h_relu0 C3} and are required for the ReLU activations of the remaining hidden layers.

We note that although not explicitly stated, the same reformulation and relaxation techniques can also be applied symmetrically to the output layer of the ANN. With the ReLU activated neurons now defined, the rest of model is identical to that defined in Proposition \ref{prop:reform_bin}. The complete MIP formulation for our problem can be found in the appendix.

\section{Greedy Layer-wise Pre-Training}

In this section, we apply our MIP formulations to the greedy layer-wise pre-training framework presented by \citep{bengio2007greedy}. Training full-scale ANNs with large data sets can be challenging even with state-of-the-art solvers despite having tight formulations \citep{toro2019training,thorbjarnarsontraining,kurtz2021efficient}. It is therefore an area of interest to explore the validity of our MIP formulation as pre-training parameters used to initialize SGD solvers. 
\begin{algorithm}[H]
\caption{Greedy Layer-wise Pre-training with a Binary MIP}\label{Greedy Binary Algorithm}
\begin{algorithmic}[1]
\Require $ L \geq 1 $, $ Y $ be input labels Y \\
Initialize $ X_{0} $ with input matrix X \\
Initialize $ h_{\ell}, \alpha_{\ell}, \beta_{\ell} = 0 $
\State $ O,W,B \gets BinaryMIP(X_{0},Y,L=1) $ 
\State $ \alpha_{0} \gets W_{0}, \beta_{0} \gets B_{0}, h_{0} \gets O_{0} $
\For{$ \ell = 1,...,L $}
 \State $ X_{\ell} \gets h_{\ell-1} $ 
 \State $ O,W,B \gets BinaryMIP(X_{\ell},Y,L=1) $ 
 \State $ \alpha_{\ell} \gets W_{0}, \beta_{\ell} \gets B_{0}, h_{\ell} \gets O_{0} $
\EndFor
\State $ \alpha_{L} \gets W_{1}, \beta_{L} \gets B_{1} $
\end{algorithmic}
\end{algorithm}

In Algorithm \ref{Greedy Binary Algorithm}, we require that we train neural networks with at least one layer, and that the data labels are constant throughout the algorithm. The input data for training layer 0 are initialized as the input matrix $ X \in \mathbbm{R}^{n \times d} $. We also define variables $ h_{\ell}, \alpha_{\ell},\beta_{\ell}$ as 0 matrices such that $ h_{\ell} = 0_{n,k}, \forall \; n \in [N]_1, k \in [K]_1, \ell \in [L-1]_0, \alpha_{0} = 0_{i,k}, \alpha_{\ell} = 0_{k,k}, \alpha_{L} = 0_{k,j}, \forall \; k \in [K]_1, j \in [J]_1, \ell \in [L-1]_1, \beta_{\ell} = 0_{k,1}, \beta_{L} = 0_{j,1} \forall \; k \in [K]_1, j \in [J]_1, \ell \in [L-1]_0 $. $ BinaryMIP(X,Y,L) $ represents a call to a function that solves the Binary MIP formulation with given input values $ X $, prediction labels $ Y $, and the number of layers $ L $ as the arguments. For Algorithm \ref{Greedy Binary Algorithm}, the number of hidden layers trained in each function call is always 1 to maintain a layer-wise training format. In the first function call, it returns the activated outputs of the $ 0^{th} $ hidden layer and output layer as matrix $ O_{0} \in \mathbbm{R}^{n \times k}, O_{1} \in \mathbbm{R}^{ n \times j} $, weight matrices $ W_{0} \in \mathbbm{R}^{d \times k}, W_{1} \in \mathbbm{R}^{k \times j} $, and bias vectors $ B_{0} \in \mathbbm{R}^{k \times 1}, B_{1} \in \mathbbm{R}^{j \times 1} $. The solved parameters from the $ 0^{th} $ layers are then saved in $ h, \alpha, \beta $. The remaining $ L-1 $ layers are trained in an iterative loop, where the input of the 1 layer neural network is initialized as the the hidden layer output of the previous iteration. Finally, the last layer parameters are saved from the final iteration's return values. 

Algorithm \ref{Greedy ReLU Algorithm}, uses a similar framework to Algorithm \ref{Greedy Binary Algorithm}, where $ ReLUMIP(X,Y,L) $ represents a call to function that solves the ReLU MIP formulation. Note that Algorithm \ref{Greedy ReLU Algorithm} differs from Algorithm \ref{Greedy Binary Algorithm} in steps 11-12, where we have another call to $ ReLUMIP $, with 0 hidden layers, in order to eliminate the need for relaxations when training the last layer.

\begin{algorithm}[H]
\caption{Greedy Layer-wise Pre-training with a ReLU MIP}\label{Greedy ReLU Algorithm}
\begin{algorithmic}[1]
\Require $ L \geq 1 $, $ Y $ be input labels Y \\
Initialize $ X_{0} $ with input matrix X \\
Initialize $ h_{\ell}, \alpha_{\ell}, \beta_{\ell} = 0 $
\State $ O,W,B \gets ReLUMIP(X_{0},Y,L=1) $ 
\State $ \alpha_{0} \gets W_{0}, \beta_{0} \gets B_{0}, h_{0} \gets O_{0} $
\For{$ \ell = 1,...,L $}
 \State $ X_{\ell} \gets h_{\ell-1} $ 
 \State $ O,W,B \gets ReLUMIP(X_{\ell},Y,L=1) $ 
 \State $ \alpha_{\ell} \gets W_{0}, \beta_{\ell} \gets B_{0}, h_{\ell} \gets O_{0} $
\EndFor
\State $ X_{L} \gets h_{L-1} $ 
\State $ O,W,B \gets ReLUMIP(X_{L},Y,L=0) $ 
\State $ \alpha_{L} \gets W_{0}, \beta_{L} \gets B_{0} $
\end{algorithmic}
\end{algorithm}

\section{Experiments and Results}

In this section, we compare our MIP based training methods against SGD and end-to-end training methods. We consider a full Binary MIP training, greedy layer-wise MIP training, greedy layer-wise MIP training as pre-training for SGD, SGD applied on binary and ReLU ANNs, greedy layer-wise SGD applied to ReLU and binary ANNs, and greedy SGD as pre-training for ReLU and binary ANNs that are then trained fully with SGD. Due to computation resources constraints, experimentation for our ReLU MIP formulation is beyond the scope of this work and will be validated in future work. We also evaluate the effectiveness of using MIP in a greedy layer-wise algorithm to obtain pre-training parameters for SGD training. The goal of these experiments is to show which methods and models can result in the most parsimonious representations that achieve a high out of sample prediction accuracy. All of the models in this section are trained using synthetic data generated from an exclusive-or (XOR) logic gate. The input data $ X \in \{x_{1}, x_{2}, x_{3}, x_{4}, x_{5}\}, x_{d} \in \{0,1\} $ produces one-hot encoded labels where an odd parity in $ x_{1}, x_{3}, x_{5} $ produces $ [0,1] $, and an even parity produces $ [1,0] $. The motivation for this structure is to test our method's ability to only select features that effect the labels. The data is randomly sampled with replacement to produce 1000 training inputs and labels, to which we then introduce $ p = 0.1 $ Bernoulli noise. The testing data was similarly obtained by randomly sampling 250 inputs and labels with replacement. Each experiment was repeated across 5 different seeds, and the average testing accuracy was used to obtain Table \ref{Table 1}. All experiments were conducted on a system configured at 1.6 GHz Dual-Core Intel Core i5 with 8 GB 1600 MHz DDR3 Memory. We used Gurobi Optimizer v9.1.2 (\cite{gurobi}) on Python 3.7.2 to solve our MIP models. 

First, we compare the various models and training methods across ANN architectures by varying the number of layers. Specifically, we train ANNs ranging from one hidden layer to five hidden layers with a fixed five units in each hidden layer. We then recorded the minimum number of hidden layers required by the combination of training methods and model to achieve 85\% out of sample accuracy to test which methods can provide the smallest effective models. SGD models were trained to 10,000 epochs. 

\begin{table}
\TABLE{List of models and minimum number of layers and units needed to train to 85\% testing accuracy\label{Table 1}}
{\begin{tabular}{|l|c|c|}
\hline
{\bfseries Model} & {\bfseries Number of Layers} & {\bfseries Number of Units } \\
\hline
Binary MIP & $ 1 $ layer & NaN\\
\hline
Greedy Binary MIP & $ 1 $ layer & $5$ units \\
\hline
Greedy Binary MIP + SGD & $ >5 $ layers & $10$ units \\
\hline
Binary SGD & $ >5 $ layers & $>50$ units \\ 
\hline
Greedy Binary SGD & $ >5 $ layers & $>50$ units \\
\hline
Greedy Binary SGD + SGD & $ >5 $ layers & $>50$ units \\
\hline
ReLU SGD & $ >5 $ layers & $50$ units \\
\hline
ReLU Greedy SGD & $ >5 $ layers & $> 50$ units\\
\hline
ReLU Greedy SGD + SGD & $ >5 $ layers & $>50$ units \\
\hline
\end{tabular}}
{}
\end{table}
As shown in Table \ref{Table 1}, the Binary MIP and Greedy Binary MIP outperform other models by requiring only one layer to achieve a test accuracy greater than our 85\% threshold. On the other hand, SGD reliant models fail to meet the threshold with the architectures chosen for this experiment. Moreover, the SGD training model fails to meet the threshold even when using our Binary MIP model solved with the greedy layer-wise algorithm as pre-training parameters. 

For the next set of experiments we fixed the number of hidden layers to 3 and varied the number of units per hidden layer. We used thresholds of 5,10,20,30,40 and 50 units and recorded the lowest threshold at which the models where able to achieve 85\% accuracy. %

Unfortunately, due to the large number of variables requires to train wide networks, our Binary MIP failed to produce any results with the computation resources at hand. However, we see in Table \ref{Table 1} that ANNs trained with the Binary MIP using the greedy layer-wise algorithm meet the required testing accuracy threshold with the fewest units chosen in the experiment. We also observed that the solver's optimal objective did not change after training the first layer, showing that the threshold was met after training just one layer at 5 units. This result shows the potential of MIP based methods in training parsimonious and accurate models. However, this does indicate that full MIP based methods may be more appropriate for narrow deep network architectures, though with a layer by layer approach MIP methods do provide an advantage over SGD based methods and perform well for pre-trainning.
\section{Conclusion}

In this paper, we proposed several MIP models to solve ANN training problems. We presented MIP formulations to train neural networks with binary activations as well as ReLU activations. To exploit the structure of the model, we also adopted greedy algorithms for our programs to train layer-by-layer. We showed that our greedy layer-wise, binary activated MIP outperforms traditional training models in two experiments. In both experiments, our model meets the testing accuracy threshold with the fewest number of layers and the fewer number of units in each layer. For architectures constrained by the number of units, we also see that our binary activated MIP is able to compete with its greedy counterpart. In essence, our models can achieve strong predictive accuracy with more parsimonious architectures compared to traditional models that require deep and neuron-dense architectures. We consider our paper to be another step in the exploration of using mixed-integer programming as a tool for deep learning problems. There are interesting challenges remaining that pertain to large-scale training problems for non-linear and non-convex structures. We see potential in future research on the application of MIP models for training ANNs that solve causal problems.

%
%
\bibliographystyle{informs2014}
\bibliography{references}


%
%
\begin{APPENDIX}{}
\section{Proofs of Propositions}

\proof{Proof of Proposition \ref{prop:nll}.}
First, we can explicitly write the negative log likelihood as: 
\begin{align}
 & -\log(\prod_{n=1}^{N}(\sigma(h_{n,j,L})^{\mathbbm{1}[y_n=j]})) = -\sum_{n=1}^{N} \mathbbm{1}[y_n=j]\log\Big(\frac{\exp(h_{n,j,L})}{\sum_{j' \in J} \exp(h_{n,j',L})}\Big) \label{NLL 1} \\
 & = -\sum_{i=n}^{N} \mathbbm{1}[y_n=j](h_{n,j.L}-\log \sum_{j' \in J} \exp(h_{n,j',L})) \label{NLL 2}
\end{align}
Here we note that the expression $\log \sum_{j' = J} \exp(h_{n,j^\prime,L}) = \Theta(\max_{j^\prime \in j} h_{n,j^\prime,L})$ (\cite{calafiore2014optimization}). Thus using this fact we obtain:
\begin{align}
 & \eqref{NLL 2} \leq -\sum_{n=1}^{N} \mathbbm{1}[y_n=j](h_{n,j,L}-\max_{j' \in J}\{h_{n,j',L}\} - \log(|J|)) \label{NLL 3} \\
 & = \sum_{n=1}^{N} \mathbbm{1}[y_n=j](\max_{j' \in J}\{h_{n,j',L}\}-h_{n,j,L} + \log(|J|)) \label{NLL 4}
\end{align}
Using the other side of the big $\Theta$ condition yields that $\eqref{NLL 2} \geq \sum_{n=1}^{N} \mathbbm{1}[y_n=j](\max_{j' \in J}\{h_{n,j',L}\}-h_{n,j,L})$. If we let $\omega_{n} = \max_{j' \in J}\{h_{n,j',L}\}$ we can use standard formulation techniques to reformulate it using linear constraints (\cite{wolsey1999integer}). In conjunction with removing the constant terms that do not depend on $\omega_n,\{h_{n,j',L}\}$ yields the desired result. 
\endproof

 \proof{Proof of Proposition \ref{prop:reform_bin}.}
 There are two key challenges for the reformulation of Constraints (2),(3),(4) first is the binary activation function itself and second are the bi-linear terms present in the hidden and output layers. Without loss of generality, let us consider a particular unit in the neural network such that $h_{n,k,\ell} = \mathbbm{1}[ \sum_{k^\prime =0}^K \alpha_{k^\prime,k,\ell}h_{n,k^\prime,\ell} + \beta_{k,\ell} \geq 0]$. First let us consider the case of the binary activation. This condition can be modeled as a disjunction (\cite{wolsey1999integer}), a form of constraint that can be reformulated as follows:
 \begin{align}
 &\sum_{k^\prime =0}^K \alpha_{k^\prime,k,\ell-1}h_{n,k^\prime,\ell} + \beta_{k,\ell} \leq M h_{n,k,\ell} \\
 & \sum_{k^\prime =0}^K \alpha_{k^\prime,k,\ell-1}h_{n,k^\prime,\ell} + \beta_{k,\ell} \geq \epsilon + (-M-\epsilon)(1- h_{n,k,\ell}) 
 \end{align}
 Where $M$ is a sufficiently large constant, and $\epsilon$ is a small constant. Next let us consider the bi-linear terms $\alpha_{k',k,\ell}h_{n,k^\prime,\ell-1}$. Here we note that $\alpha_{k',k,\ell} \in [\alpha^L,\alpha^U]$ is a continuous real-valued variable, while $h_{n,k^\prime,\ell-1} \in \mathbbm{B}$ is binary valued. This type of products can be reformulated through a standard technique by introducing an additional variable $z_{n,k^\prime,k,\ell} = \alpha_{k^\prime,k,\ell}h_{n,k^\prime,\ell-1}$. Then these products can be written as:
 \begin{align}
 & z_{n,k^\prime,k,\ell} \le \alpha_{k^\prime,k,\ell} + M(1-h_{n,k^\prime,\ell-1}) \\
 & z_{n,k^\prime,k,\ell} \ge \alpha_{k^\prime,k,\ell} -M(1-h_{n,k^\prime,\ell-1}) \\ 
 & -Mh_{n,k^\prime,\ell-1} \le z_{n,k^\prime,k,\ell} \le Mh_{n,k^\prime,\ell-1} 
 \end{align}
 Where again $M$ is an appropriately picked sufficiently large constant. Using these reformulations where appropriate and making the necessary variable substitutions results in the formulations presented above.
 \endproof
 
\proof{Proof of Proposition \ref{prop:reform_relu}.}
Much like the proof of Proposition 2 the two main challenges for reformulation involve the piece-wise nature of the activation and reformulation of the bi-linear terms. Without loss of generality consider a single unit evaluated at a particular data point with specific indices $\ell,k,n$. If we use the definition $h_{n,k,\ell} = \mathbbm{1}[ \sum_{k^\prime =0}^K \alpha_{k^\prime,k,\ell}h_{n,k^\prime,\ell} + \beta_{k,\ell} \geq 0]$ then we can rewrite the ReLU activation conditions as:
\begin{equation}
 h^{ReLU}_{n,k,\ell} = 
 \begin{cases}
 & \alpha_{k^\prime,k,\ell}h^{ReLU}_{n,k,\ell-1} + \beta_{k,\ell}, \textrm{ if }h_{n,k,\ell} = 1 \\
 & 0, \textrm{ if } h_{n,k,\ell}= 0
 \end{cases}
\end{equation}
Hence we can use the disjunction constraint reformulation from Proposition 2 to obtain the conditions that ensure $h_{n,k,\ell}$ when appropriate. Using disjunctions (\cite{wolsey1999integer}) again on the conditions of when $h^{ReLU}_{n,k,\ell}$ is equal to zero and using $h_{n,k,\ell}$ as the binary indicator we can model the conditions above resulting in the desired constraints. For the bilinear terms, we introduce the proper auxiliary variables and bounds as proposed in (\cite{westerlund2011convex}) using general disjunctive programming (\cite{raman1994modelling}).
\endproof

\section{Mixed-Integer Programming Formulation with Binary Activated Neural Networks}

\begin{table}[H]
\TABLE
{A summary of parameters and decision varables used in the binary model \label{Binary MIP Parameters}}
{\begin{tabular}{p{2cm} p{7cm} p{1.15cm}} 
 \hline
 \textbf{Parameter} & \textbf{Description} & \textbf{Range} \\
 \hline
 $ x_{n,i} $ & Vector inputs of size n$ \times $i, where $ n \in [N]_1 $ and $ i \in [d]_1 $. N is the number of data points, d is the number of dimensions/features. & $ [-\infty, \infty] $ \\
 $ y_{n,j} $ & Binary vector outputs of size n$ \times $j indicating category selection, where $ n \in [N]_1 $ and $ j \in [J]_1 $. N is the number of data points and J is the dimension of the label. & $ \{0,1\} $\\
 $ \alpha_{i,k,0} $ & Weight for feature i in unit k in the $ 0^{th} $ hidden layer, $ \forall \ i\in [d]_1,k \in [K]_1. $ & $ [\alpha^{L}, \alpha^{U}] $ \\
 $ \alpha_{k^\prime,k,\ell} $ & Weight from the $ k\prime^{th} $ unit in layer $\ell-1$ to the $ k^{th} $ in layer $\ell$, $ \forall\ k^\prime,\ k \in [K]^{2}_{1},\ell \in [L-1]_1 $. & $ [\alpha^{L}, \alpha^{U}] $ \\
 $ \alpha_{k^\prime,j,L} $ & Weight from the $ k\prime^{th} $ unit in hidden layer $L-1$ to the $ j^{th} $ unit in the output layer $L$, $ \forall\ \ k^\prime \in [K]_1,j \in [J]_1 $. & $ [\alpha^{L}, \alpha^{U}] $ \\
 $ \beta_{k,\ell} $ & Bias for unit k in layer $\ell$, $ \forall \ k \in [K]_1,\ell \in [L-1]_0 $. & $ [\beta^{L}, \beta^{U}] $ \\
 $ \beta_{j,L} $ & Bias for unit j in the final layer, $ \forall \ j \in [J]_1 $. & $ [\beta^{L}, \beta^{U}] $\\
 $ h_{n,k,\ell} $ & Binary output of unit k in layer $\ell$, $ \forall\ n \in [N]_1,k \in [K]_1,\ell \in [L-1]_0 $. & $ \{0,1\} $\\
 $ h_{n,j,L} $ & Output of final layer, $ \forall\ n \in [N]_1,\ j \in [J]_1 $. & $ [-\infty, \infty] $ \\
 $ \omega_{n} $ & Placeholder variable for $ \max_{j\in J}\{h_{n,j,L}\}, \forall \ n \in [N]_1 $. & $ [-\infty, \infty] $ \\
 $ r_{n,j,j^\prime}: $ & Binary helper variable to diversify output layer of $ h_{n,j,L}, \forall \ n \in [N]_1, \ j,j^\prime \in [J]^{2}_{1}, \ j \neq j^\prime $. & $ \{0,1\} $ \\
 $ z_{n,k^\prime,k,\ell} $ & Auxiliary variable that represents $ \alpha_{k^\prime,k,\ell}h_{n,k^\prime,\ell-1}, \forall \ n \in [N]_1, k^\prime,k \in [K]^{2}_{1},\ell \in [L-1]_1 $. & $ [\alpha^{L},\alpha^{U}] $ \\
 $ z_{n,k^\prime,j,L} $ & Auxiliary variable that represents $ \alpha_{k^\prime,j,L}h_{n,k^\prime,L-1}, \forall \ n \in [N]_1,k^\prime \in [K]_1,j \in [J]_1 $. & $ [\alpha^{L},\alpha^{U}] $ \\
 \hline
\end{tabular}}
{}
\end{table}

\textbf{Formulation:}

\begin{flalign}
 \min_{\alpha,\beta,h,z,\omega,r} \sum_{n=1}^{N} \sum_{j=1}^{J} y_{n,j}(\omega_{n}-h_{n,j,L})&& 
\end{flalign}
subject to \\
For all $k \in [K]_1,n \in [N]_1$:
\begin{align}
 & \sum_{i=1}^{d} (\alpha_{i,k,0}x_{n,i}) + \beta_{k,0} \le Mh_{n,k,0} \label{binary_formulation h0_define C1} \\
 & \sum_{i=1}^{d} (\alpha_{i,k,0}x_{n,i}) + \beta_{k,0} \ge \epsilon + (-M-\epsilon)(1-h_{n,k,0}) \label{binary_formulation h0_define C2}
\end{align}
For all $\ell \in [L-1]_1, k \in [K]_1, n \in [N]_1$:
\begin{align}
 & \sum_{k^\prime=1}^{K} (z_{n,k^\prime,k,\ell}) + \beta_{k,\ell} \le Mh_{n,k,\ell} \label{binary_formulation hl_define C1} \\
 & \sum_{k^\prime=1}^{K} (z_{n,k^\prime,k,\ell}) + \beta_{k,\ell} \ge \epsilon + (-M-\epsilon)(1-h_{n,k,\ell}) \label{binary_formulation hl_define C2} \\
 & z_{n,k^\prime,k,\ell} \le \alpha_{k^\prime,k,\ell} + M(1-h_{n,k^\prime,\ell-1}) \label{binary_formulation zl_define C1} \\
 & z_{n,k^\prime,k,\ell} \ge \alpha_{k^\prime,k,\ell} -M(1-h_{n,k^\prime,\ell-1}) \label{binary_formulation zl_define C2} \\ 
 & -Mh_{n,k^\prime,\ell-1} \le z_{n,k^\prime,k,\ell} \le Mh_{n,k^\prime,\ell-1} \label{binary_formulation zl_define C3}
\end{align} 
For all $j \in [J]_1, n \in [N]_1$:
\begin{align}
 & h_{n,j,L} \le \sum_{k^\prime=0}^{K} (z_{n,k^\prime,j,L}) + \beta_{j,L} \label{binary_formulation hL_define C1} \\
 & h_{n,j,L} \ge \sum_{k^\prime=0}^{K} (z_{n,k^\prime,j,L}) + \beta_{j,L} \label{binary_formulation hL_define C2} \\
 & z_{n,k^\prime,j,L} \le \alpha_{k^\prime,j,L} + M(1-h_{n,k^\prime,L-1}) \label{binary_formulation zL_define C1} \\
 & z_{n,k^\prime,j,L} \ge \alpha_{k^\prime,j,L} - M(1-h_{n,k^\prime,L-1}) \label{binary_formulation zL_define C2} \\
 & -Mh_{n,k^\prime,L-1} \le z_{n,k^\prime,j,L} \le Mh_{n,k^\prime,L-1} \label{binary_formulation zL_define C3} \\
 & \omega_{n} \ge h_{n,j,L} \label{binary_formulation omega_define} \\
 & h_{n,j,L} + h_{n,j^\prime,L} - 2h_{n,j,L} \leq - \epsilon + Mr_{n,j,j^\prime} \ \forall \ j,j^\prime \in [J]^{2}_{1}, j \neq j^\prime \label{binary_formulation diversify C1} \\ 
 & h_{n,j,L} + h_{n,j^\prime,L} - 2h_{n,j,L} \geq \epsilon - M(1-r_{n,j,j^\prime}) \ \forall \ j,j^\prime \in [J]^{2}_{1}, j \neq j^\prime \label{binary_formulation diversify C2}
\end{align}

\section{Mixed-Integer Programming Formulation with ReLU Activated Neural Networks}
\label{sec:full_relu}
\begin{table}[H]
\TABLE
{A summary of parameters used in the ReLU model \label{ReLU MIP Parameters}}
{\begin{tabular}{p{2cm} p{7cm} p{1.15cm}} 
 \hline
 \textbf{Parameter} & \textbf{Description} & \textbf{Range} \\
 \hline
 $ x_{n,i} $ & Vector inputs of size n$ \times $i, where $ n \in [N]_1 $ and $ i \in [d]_1 $. N is the number of data points, d is the number of dimensions/features. & $ [-\infty, \infty] $ \\
 $ y_{n,j} $ & Binary vector outputs of size n$ \times $j indicating category selection, where $ n \in [N]_1 $ and $ j \in [J]_1 $. N is the number of data points and J is the dimension of the label. & $ \{0,1\} $\\
 \hline
\end{tabular}}
{}
\end{table}

\begin{table}[H]
\TABLE
{A summary of decision variables used in the ReLU model \label{ReLU MIP Variables}}
{\begin{tabular}{p{2cm} p{7cm} p{1.15cm}} 
 \hline
 \textbf{Variable} & \textbf{Description} & \textbf{Range} \\
 \hline
 $ \alpha_{i,k,0} $ & Weight for feature i in unit k in the $ 0^{th} $ hidden layer, $ \forall i \in [d]_1,k \in [K]_1 $. & $ [\alpha^{L}, \alpha^{U}] $ \\
 $ \alpha_{k^\prime,k,\ell} $ & Weight from the $ k\prime^{th} $ unit in layer $\ell-1$ to the $ k^{th} $ unit in layer $\ell$, $ \forall\ k^\prime,k \in [K]^{2}_{1},\ell \in [L-1]_1 $. & $ [\alpha^{L}, \alpha^{U}] $ \\
 $ \alpha_{k^\prime,j,L} $ & Weight from the $ k\prime^{th} $ unit in the $ (L-1)^{st} $ layer to the $j^{th}$ in the $ L^{th} $, or output, layer, $ \forall\ k^\prime \in [K]_1,j \in [J]_1 $. & $ [\alpha^{L}, \alpha^{U}] $ \\
 $ \beta_{k,\ell} $ & Bias for unit k in layer $\ell$, $ \forall\ k \in [K]_1,\ell\in [L-1]_1 $. & $ [\beta^{L}, \beta^{U}] $ \\
 $ \beta_{j,L} $ & Bias for unit j in the output layer, L, $ \forall j \in [J]_1 $. & $ [\beta^{L}, \beta^{U}] $ \\
 $ h_{n,k,\ell} $ & Binary output of unit k in the $ \ell^{th} $ layer, $ \forall n \in [N]_1,k \in [K]_1, \ \ell \in [L-1]_0 $ & $ \{0,1\} $ \\
 $ h^{ReLU}_{n,k,\ell} $ & ReLU activated output of unit k in the $ \ell^{th} $ layer, $ \forall\ n \in [N]_1,k \in [K]_1,\ell \in [L-1]_0 $. & $ [0,\infty] $ \\
 $ h_{n,j,L} $ & Output of final layer, $ \forall\ n \in [N]_1,j \in [J]_1 $. & $ [-\infty, \infty] $ \\
 $ \omega_{n} $ & Placeholder variable for $ \max_{j\in J}\{h_{n,j,L}\} \ \forall \ n \in [N]_1 $. & $ [-\infty, \infty] $ \\
 $ r_{n,j,j^\prime} $ & Binary helper variable to diversify output layer of $ h_{n,j,L} \forall \ n \in [N]_1, j,j^\prime \in [J]^{2}_{1}, \ j \neq j^\prime $. & $ \{0,1\} $ \\
 $ z_{n,k^\prime,k,\ell} $ & Auxiliary variable that represents $ \alpha_{k^\prime,k,\ell}h^{ReLU}_{n,k^\prime,\ell-1} \ \forall \ n \in [N]_1, k^\prime,k \in [K]^{2}_{1},\ell \in [L-1]_1 $. & $ [0,\infty] $ \\
 $ z_{n,k^\prime,j,L} $ & Auxiliary variable that represents $ \alpha_{k^\prime,j,L}h^{ReLU}_{n,k^\prime,L-1} \ \forall \ n \in [N]_1, \ k^\prime \in [K]_1, \ j \in [J]_1 $. & $ [0,\alpha^{U}] $ \\
 $ \lambda_{k^\prime,k,\ell,p} $ & Binary variable that indicates which partitions p of the McCormick relaxation needs to be active, where we have P partitions for the hidden layers $ \forall \ k,k^\prime \in [K]^{2}_{1},\ell \in [L]_1, \ p \in [P]_1 $. & $ \{0,1\} $ \\
 $ \lambda_{k^\prime,j,L,p} $ & Binary variable that indicates which partition of the McCormick relaxation needs to be active, where we have P partitions for the output layer $ \forall \ k^\prime \in [K]_1, \ j \in [J]_1, \ p \in [P]_1 $. & $ \{0,1\} $ \\
 \hline
\end{tabular}}
{}
\end{table}

\textbf{Formulation:}
\begin{flalign}
 \min_{\alpha,\beta,h,h^{ReLU},\omega,r, \lambda} \sum_{n=1}^{N} \sum_{j=1}^{J} y_{n,j}(\omega_{n}-h_{n,j,L})&& 
\end{flalign} 
subject to \\
For all $ k \in [K]_1, n \in [N]_1 $
\begin{align}
 & \sum_{i=1}^{d} (\alpha_{i,k,0}x_{n,i}) + \beta_{k,0} \le Mh_{n,k,0} \label{relu_formulation h0_define C1} \\
 & \sum_{i=1}^{d} (\alpha_{i,k,0}x_{n,i}) + \beta_{k,0} \ge \epsilon + (-M-\epsilon)(1-h_{n,k,0}) \label{relu_formulation h0_define C2} \\
 & h^{ReLU}_{n,k,0} \le (\sum_{i=0}^{D} (\alpha_{i,k,0}x_{n,i}) + \beta_{k,0}) + M(1-h_{n,k,0}) \label{relu_formulation h_relu0_define C1} \\
 & h^{ReLU}_{n,k,0} \ge (\sum_{i=0}^{D} (\alpha_{i,k,0}x_{n,i}) + \beta_{k,0}) - M(1-h_{n,k,0}) \label{relu_formulation h_relu0_define C2} \\
 & -Mh_{n,k,0} \le h^{ReLU}_{n,k,0} \le Mh_{n,k,0} \label{relu_formulation h_relu0_define C3}
\end{align}
For all $ \ell \in [L-1]_1, k \in [K]_1, n \in [N]_1 $
\begin{align}
 & \sum_{k^\prime=1}^{K} (z_{n,k^\prime,k,\ell}) + \beta_{k,\ell} \le Mh_{n,k,\ell} \label{relu_formulation hl_define C1} \\
 & \sum_{k^\prime=1}^{K} (z_{n,k^\prime,k,\ell}) + \beta_{k,\ell} \ge \epsilon + (-M-\epsilon)(1-h_{n,k,\ell}) \label{relu_formulation hl_define C2} \\
 & h^{ReLU}_{n,k,\ell} \le (\sum_{k^\prime=1}^{K} (z_{n,k^\prime,k,\ell}) + \beta_{k,\ell}) + M(1-h_{n,k,\ell}) \label{relu_formulation h_relul_define C1} \\
 & h^{ReLU}_{n,k,\ell} \ge (\sum_{k^\prime=1}^{K} (z_{n,k^\prime,k,\ell}) + \beta_{k,\ell}) - M(1-h_{n,k,\ell}) \label{relu_formulation h_relul_define C2} \\
 & -Mh_{n,k,\ell} \le h^{ReLU}_{n,k,\ell} \le Mh_{n,k,\ell} \label{relu_formulation h_relul_define C3} 
\end{align}
For all $ p \in [P]_1, \ell \in [L-1]_1, k,k^\prime \in [K]^{2}_{1}, n \in [N]_1 $
\begin{align}
 & z_{n,k^\prime,k,\ell} \geq \alpha_{p}^{L}h^{ReLU}_{n,k^\prime,\ell-1} + \alpha_{k^\prime,k,\ell}h_{ReLU}^{L} - \alpha_{p}^{L}h_{ReLU}^{L} - M(1-\lambda_{k^\prime,k,\ell,p}) \label{relu_formulation z_l McCormick C1} \\
 & z_{n,k^\prime,k,\ell} \geq \alpha_{p}^{U}h^{ReLU}_{n,k^\prime,\ell-1} + \alpha_{k^\prime,k,\ell}h_{ReLU}^{U} - \alpha_{p}^{U}h_{ReLU}^{U} - M(1-\lambda_{k^\prime,k,\ell,p}) \label{relu_formulation z_l McCormick C2} \\
 & z_{n,k^\prime,k,\ell} \leq \alpha_{p}^{U}h^{ReLU}_{n,k^\prime,\ell-1} + \alpha_{k^\prime,k,\ell}h_{ReLU}^{L} - \alpha_{p}^{U}h_{ReLU}^{L} + M(1-\lambda_{k^\prime,k,\ell,p}) \label{relu_formulation z_l McCormick C3} \\
 & z_{n,k^\prime,k,\ell} \leq \alpha_{p}^{L}h^{ReLU}_{n,k^\prime,\ell-1} + \alpha_{k^\prime,k,\ell}h_{ReLU}^{U} - \alpha_{p}^{L}h_{ReLU}^{U} + M(1-\lambda_{k^\prime,k,\ell,p}) \label{relu_formulation z_l McCormick C4} 
\end{align}
For all $ \ell \in [L-1]_1, k,k\prime \in [K]^{2}_{1} $
\begin{align}
 & \sum_{p=1}^{P}\alpha_{p}^{L}\lambda_{k^\prime,k,\ell,p} \le \alpha_{k^\prime,k,\ell} \le \sum_{p=1}^{P}\alpha_{p}^{U}\lambda_{k^\prime,k,\ell,p} \label{relu_formulation alpha_l_bound} \\
 & \sum_{p=0}^{P} \lambda_{k^\prime,k,\ell,p} = 1 \label{relu_formulation lambda_l} 
\end{align}
For all $ j \in [J]_1, n \in [N]_1 $
\begin{align}
 & h_{n,j,L} \le \sum_{k^\prime=0}^{K} (z_{n,k^\prime,j,L}) + \beta_{j,L} \label{relu_formulation h_L define C1} \\
 & h_{n,j,L} \ge \sum_{k^\prime=0}^{K} (z_{n,k^\prime,j,L}) + \beta_{j,L} \label{relu_formulation h_L define C2}
\end{align}
For all $ p \in [P]_1, j \in [J]_1, k^\prime \in [K]_1, n \in [N]_1 $ 
\begin{align}
 & z_{n,k^\prime,j,L} \geq \alpha_{p}^{L}h^{ReLU}_{n,k^\prime,L-1} - M(1-\lambda_{k^\prime,j,L,p}) \label{relu_formulation z_L McCormick C1} \\
 & z_{n,k^\prime,j,L} \geq \alpha_{p}^{U}h^{ReLU}_{n,k^\prime,L-1} + \alpha_{k^\prime,j,L}h_{ReLU}^{U} - \alpha_{p}^{U}h_{ReLU}^{U} - M(1-\lambda_{k^\prime,j,L,p}) \label{relu_formulation z_L McCormick C2} \\
 & z_{n,k^\prime,j,L} \leq \alpha_{p}^{U}h^{ReLU}_{n,k^\prime,L-1} + M(1-\lambda_{k^\prime,j,L,p}) \label{relu_formulation z_L McCormick C3} \\
 & z_{n,k^\prime,j,L} \leq \alpha_{p}^{L}h^{ReLU}_{n,k^\prime,L-1} + \alpha_{k^\prime,j,L}h_{ReLU}^{U} - \alpha_{p}^{L}h_{ReLU}^{U} + M(1-\lambda_{k^\prime,j,L,p}) \label{relu_formulation z_L McCormick C4} 
\end{align}
For all $ j \in [J]_1, k^\prime \in [K]_1 $
\begin{align}
 & \sum_{p=1}^{P}\alpha_{p}^{L}\lambda_{k^\prime,j,L,p} \le \alpha_{k^\prime,j,L} \le \sum_{p=1}^{P}\alpha_{p}^{U}\lambda_{k^\prime,j,L,p} \label{relu_formulation alpha_L_bound} \\
 & \sum_{p=0}^{P} \lambda_{k^\prime,j,L,p} = 1 \label{relu_formulation lambda_L}
\end{align}
For all $ j \in [J]_1, n \in [N]_1 $
\begin{align}
 & \omega_{n} \ge h_{n,j,L} \ \forall n \in N, \ j \in J \label{relu_formulation omega_define}
\end{align}
For all $ j \in [J]_1, n \in [N]_1 $
\begin{align}
 & h_{n,i,L} + h_{n,j,L} - 2h_{n,i,L} \leq - \epsilon + Mr_{n,j,j^\prime} \ \forall j,j^\prime \in [J]^{2}_{1}, j \neq j^\prime \label{relu_formulation diversify C1} \\
 & h_{n,i,L} + h_{n,j,L} - 2h_{n,i,L} \geq \epsilon - M(1-r_{n,j,j^\prime}) \ \forall j,j^\prime \in [J]^{2}_{1}, j \neq j^\prime \label{relu_formulation diversify C2} 
\end{align}

\section{Output Layer Formulation}

\begin{table}[H]
\TABLE
{A summary of parameters and decision variables used in the output layer model \label{Output Layer MIP Parameters}}
{\begin{tabular}{p{2cm} p{7cm} p{1.15cm}} 
 \hline
 \textbf{Parameter} & \textbf{Description} & \textbf{Range} \\
 \hline
 $ x_{n,i} $ & Vector inputs of size n$ \times $i, where $ n \in [N]_1 $ and $ i \in [d]_1 $. N is the number of data points, d is the number of dimensions/features. & $ [-\infty, \infty] $ \\
 $ y_{n,j} $ & Binary vector outputs of size n$ \times $j indicating category selection, where $ n \in [N]_1 $ and $ j \in [J]_1 $. N is the number of data points and J is the dimension of the label. & $ \{0,1\} $\\
 $ \alpha_{j,j^\prime,0} $ & Weight for feature i in unit j in the output layer, $ \forall\ i\in [d]_1,\ j \in [J]_1 $. & $ [\alpha^{L}, \alpha^{U}] $ \\
 $ h_{n,j,0} $ & Output of unit j in the output layer, $ \forall\ n \in [N]_1, \ j \in [J]_1 $. & $ [-\infty,\infty] $\\
 $ \omega_{n} $ & Placeholder variable for $ \max_{j\in J}\{h_{n,j,0}\} \ \forall \ n \in [N]_1 $. & $ [-\infty, \infty] $ \\
 $ r_{n,j,j^\prime}: $ & Binary helper variable to diversify output layer of $ h_{n,j,0} \ \forall \ n \in [N]_1, \ j,j^\prime \in [J]^{2}_{1}, \ j \neq j^\prime $. & $ \{0,1\} $ \\
 \hline
\end{tabular}}
{}
\end{table}

\textbf{Formulation:}

\begin{flalign}
 \min_{\alpha,\beta,h,\omega,r} \sum_{n=1}^{N} \sum_{j=1}^{J} y_{n,j}(\omega_{n}-h_{n,j,0}) &&
\end{flalign}
subject to \\
For all $ j \in [J]_1, n \in [N]_1 $
\begin{align}
 & \sum_{d=0}^{D} (\alpha_{d,j,0}x_{n,d}) + \beta_{j,0} \le h_{n,j,0} \label{output_formulation h0_define C1} \\
 & \sum_{d=0}^{D} (\alpha_{d,j,0}x_{n,d}) + \beta_{k,0} \ge h_{n,j,0} \label{output_formulation h0_define C2} \\
 & \omega_{n} \ge h_{n,j,0} \label{output_formulation omega_define}
\end{align}
For all $ n \in [N]_1 $
\begin{align}
 & h_{n,i,0} + h_{n,j,0} - 2h_{n,i,0} \leq - \epsilon +
 Mr_{n,j,j^\prime}, \forall j,j^\prime \in [J]^{2}_{1}, j \neq j^\prime \label{output_formulation diversification C1} \\
 & h_{n,i,0} + h_{n,j,0} - 2h_{n,i,0} \geq \epsilon - M(1-r_{n,j,j^\prime}), \forall j,j^\prime \in [J]^{2}_{1}, j \neq j^\prime \label{output_formulation diversification C2}
\end{align}
\end{APPENDIX}
\end{document}